\documentclass[letterpaper, 10 pt, journal, twoside]{IEEEtran}
\usepackage{hyperref}
\usepackage{amsmath}
\usepackage{amssymb}
\usepackage{algorithm}
\usepackage{algpseudocode}
\usepackage{graphicx}
\usepackage{multirow} 
\usepackage{nicematrix}
\newcolumntype{C}{>{\centering\arraybackslash}X}
\usepackage{acronym}
\usepackage{cleveref}
\usepackage{subcaption}
\usepackage{siunitx}
\usepackage{svg}
\usepackage{url}
\usepackage{todonotes}
\usepackage{cite}
\usepackage{pgf} % Required for calculations
\usepackage{algorithm}
\usepackage{algpseudocode}
\usepackage{bm}       % for bold symbols
\usepackage{adjustbox}

\usepackage{graphicx}
\usepackage{overpic}
\usepackage{xcolor}

\usepackage{cleveref}

\crefname{figure}{Fig.}{Figs.}
\Crefname{figure}{Fig.}{Figs.}

\usepackage[utf8]{inputenc}  % Make sure your encoding is set to UTF-8
\DeclareUnicodeCharacter{2212}{-}

\acrodef{bim}[BIM]{Building Information Modeling}
\acrodef{vjm}[VJM]{Virtual Joint Method}

\acrodef{nocalib}[NC]{No Calibration}
\acrodef{armcalib}[AC]{Arm Calibration}
\acrodef{deflmodel}[DM]{Deflection Model}
\acrodef{baseaccel}[BA]{Base Accel.}
\acrodef{colaccel}[CA]{Column Accel.}

\newcommand{\kinematicchain}{\{\mathbf{T}_{\text{chain}}^{(t)}\}}
\newcommand{\armcalibration}{\{\mathbf{T}_{\text{arm}_{i,j}}\}}
\newcommand{\prodoverkinchain}{\prod_{\textbf{T}_{i,j} \in \kinematicchain} \!\!\!\!\!\!\!\! \mathbf{T}_{i,j}}

\newcommand{\SOthree}{\text{SO}(3)}
\newcommand{\SEthree}{\text{SE}(3)}
\newcommand{\RNbyfourbyfour}{\mathbb{R}^{N \times 4 \times 4}}

\newcommand{\Rthree}{\mathbb{R}^{3}}
\newcommand{\Rsix}{\mathbb{R}^{6}}

\newcommand{\extractpart}[2]{\left[ #1\right]_{#2}}
\newcommand{\pospart}[1]{\extractpart{#1}{\mathbf{t}}}
\newcommand{\rotpart}[1]{\extractpart{#1}{\mathbf{R}}}

\DeclareMathOperator{\foptimize}{optimize}
\DeclareMathOperator{\fcalibrate}{calibrate}
\DeclareMathOperator{\fmoveToAndRecordPoint}{moveToAndRecordPoint}
\DeclareMathOperator{\fmoveToAndRecordPoints}{moveToAndRecordPoints}
\DeclareMathOperator{\fexecuteTask}{executeTask}

\acrodef{vjmbt}[VJM\&BT]{VJM \& Base Tilt}
\acrodef{fg}[FG]{Factor Graph}
\newcommand{\tableconfigVJM}{VJM}
\newcommand{\tableconfigVJMBA}{VJM\&BT}
\newcommand{\tableconfigFULL}{FG (ours)}

\newcommand{\dnFlat}{Flat}
\newcommand{\dnOrthoWood}{Orthogonal Wood}
\newcommand{\dnDiagWood}{Diagonal Wood}
\newcommand{\dnLeftTrackWood}{Wood Left}
\newcommand{\dnSeesaw}{Seesaw}
\newcommand{\dnPallet}{Pallet}
\newcommand{\dnOutdoor}{Outdoor}

\newcommand{\percdiff}[2]{%
  \pgfmathparse{100*abs(1-(#1/#2))}%
  $\pgfmathprintnumber[precision=0]{\pgfmathresult}\%$ ($#2\si{mm}$ to $#1\si{mm}$)}

\newcommand{\capreductionfigures}{-7pt}
\newcommand{\capreductiontables}{-7pt}
\newcommand{\additionaltopvspacetables}{5pt}
\newcommand{\additionaltopvspacefigures}{1pt}
\newcommand{\topvspacealgorithm}{-9pt}

\usepackage{xcolor}
\usepackage{listings}  
\begin{document}

% -- COVER PAGE --
\begin{titlepage}
  \onecolumn
  \vspace*{2cm}
  \begin{center}
    {\Large \bfseries Preprint Version}\\[1em]

{\bfseries Enhancing Robotic Precision in Construction: \\
A Modular Factor Graph-Based Framework to Deflection and \\
Backlash Compensation Using High-Accuracy Accelerometers}\\[1em]

\textit{Accepted in IEEE Robotics and Automation Letters on November 2024}

\vspace{2em}

% -- IEEE Notice (Smaller text) --
\begin{minipage}{0.95\textwidth}
\small
\textcopyright 2024 IEEE.  Personal use of this material is permitted.  Permission from IEEE must be obtained for all other uses, in any current or future media, including reprinting/republishing this material for advertising or promotional purposes, creating new collective works, for resale or redistribution to servers or lists, or reuse of any copyrighted component of this work in other works.\\
\textbf{DOI:} \href{https://doi.org/10.1109/LRA.2024.3506276}{10.1109/LRA.2024.3506276}
\end{minipage}

\vspace{2em}

% -- Citation suggestion --
\hrule
\vspace{1em}
\textbf{Please cite this publication as:}
\vspace{0.5em}

\begin{minipage}{0.95\textwidth}
\small
J.~Kindle, M.~Loetscher, A.~Alessandretti, C.~Cadena and M.~Hutter, ''\textit{Enhancing Robotic Precision in Construction: A Modular Factor Graph-Based Framework to Deflection and Backlash Compensation Using High-Accuracy Accelerometers}'' in IEEE Robotics and Automation Letters, vol.~10, no.~1, pp.~288-295, Jan.~2025, doi:~10.1109/LRA.2024.3506276.
\end{minipage}
\vspace{1em}
\hrule

\vspace{2em}

\textbf{Bib\TeX~entry:}\\[0.5em]
\begin{lstlisting}
@ARTICLE{kindle_enhancing,
    author={Kindle, Julien and Loetscher, Michael and Alessandretti, Andrea and Cadena, Cesar and Hutter, Marco},
    journal={IEEE Robotics and Automation Letters},
    title={Enhancing {Robotic} {Precision} in {Construction}: A {Modular} {Factor} {Graph}-{Based} {Framework} to {Deflection} and {Backlash} {Compensation} using {High}-{Accuracy} {Accelerometers}},
    year={2025},
    volume={10},
    number={1},
    pages={288-295},
    keywords={Robots;Accuracy;Kinematics;Accelerometers;Robot sensing systems;Position measurement;Robot kinematics;End effectors;Measurement uncertainty;Calibration;Robotics and automation in construction;localization;sensor fusion},
    doi={10.1109/LRA.2024.3506276}
}
\end{lstlisting}

\end{center}
  \vfill
\end{titlepage}

\twocolumn  % revert to normal IEEE 2-column layout

\title{ %\LARGE \bf
Enhancing Robotic Precision in Construction: A Modular Factor Graph-based Framework to Deflection and Backlash Compensation Using High-Accuracy Accelerometers}

\author{Julien Kindle$^{1,2}$, Michael Loetscher$^{2}$, Andrea Alessandretti$^{2}$, Cesar Cadena$^{1}$ and Marco Hutter$^{1}$% <-this % stops a space
\thanks{Manuscript received: July 23, 2024; Revised: October 14, 2024; Accepted: November 18, 2024.}%Use only for final RAL version
\thanks{This paper was recommended for publication by Editor Sven Behnke upon evaluation of the Associate Editor and Reviewers' comments.}
\thanks{This work was supported by Hilti AG, Schaan, Liechtenstein.}%
\thanks{$^{1}$Julien Kindle, Cesar Cadena and Marco Hutter are with the Robotic Systems Lab, ETH Zurich, 8092 Zurich, Switzerland {\tt\footnotesize jkindle@ethz.ch}, {\tt\footnotesize cesarc@ethz.ch}, {\tt\footnotesize mahutter@ethz.ch}}%
\thanks{$^{2}$Julien Kindle, Michael Loetscher, Andrea Alessandretti are with Hilti AG, 9494 Schaan, Liechtenstein {\tt\footnotesize kindjul@hilti.com}, {\tt\footnotesize michloetscher@gmail.com}, {\tt\footnotesize alesand@hilti.com}}%
\thanks{Digital Object Identifier (DOI): see top of this page.}
}

% The paper headers
\markboth{IEEE Robotics and Automation Letters. Preprint Version. Accepted November, 2024}
{Kindle \MakeLowercase{\textit{et al.}}: Enhancing Robotic Precision in Construction}

\maketitle
% \thispagestyle{empty}
% \pagestyle{empty}

%%%%%%%%%%%%%%%%%%%%%%%%%%%%%%%%%%%%%%%%%%%%%%%%%%%%%%%%%%%%%%%%%%%%%%%%%%%%%%%%
\begin{abstract}
Accurate positioning is crucial in the construction industry, where labor shortages highlight the need for automation. Robotic systems with long kinematic chains are required to reach complex workspaces, including floors, walls, and ceilings. These requirements significantly impact positioning accuracy due to effects such as deflection and backlash in various parts along the kinematic chain. In this work, we introduce a novel approach that integrates deflection and backlash compensation models with high-accuracy accelerometers, significantly enhancing position accuracy. Our method employs a modular framework based on a factor graph formulation to estimate the state of the kinematic chain, leveraging acceleration measurements to inform the model. Extensive testing on publicly released datasets, reflecting real-world construction disturbances, demonstrates the advantages of our approach. The proposed method reduces the $95\%$ error threshold in the xy-plane by $50\%$ compared to the state-of-the-art Virtual Joint Method, and by $31\%$ when incorporating base tilt compensation.
\end{abstract}


%%%%%%%%%%%%%%%%%%%%%%%%%%%%%%%%%%%%%%%%%%%%%%%%%%%%%%%%%%%%%%%%%%%%%%%%%%%%%%%%
\section{INTRODUCTION}
\IEEEPARstart{T}{he} automation of labor-intensive tasks is crucial for many sectors, particularly the construction industry, which faces a significant labor shortage. McKinsey reports 400,000 unfilled positions in the US construction industry as of October 2021~\cite{mckinsey2024}, with this trend expected to continue.

Accurate positioning is a crucial element in the automation of many applications in the construction industry, with mechanical, electrical, and plumbing installations requiring subcentimeter accuracy relative to a building coordinate system. This is especially challenging considering robotic arms with long kinematic chains necessary to access the complex workspace of construction sites, including floors, walls, and ceilings, due to their susceptibility to model inaccuracies arising from deflections, backlash, temperature variations, and wear.

One common solution is the use of total stations, advanced surveying instruments that measure angles and distances to determine precise three-dimensional positions. By capturing horizontal and vertical angles along with the distance to a reflective target, total stations effectively utilize spherical coordinate measurements to calculate positions with millimeter-level accuracy within the building coordinate system. However, implementing closed-loop positioning~\cite{lanegger2023chasing} of a robot's end effector using total stations is often infeasible. This is due to the constant need for a clear line of sight between the total station and the tracked reflector, as well as limitations imposed by the tracking algorithm's velocity and time synchronization requirements.

\begin{figure}[t]
\includegraphics[width=\linewidth]{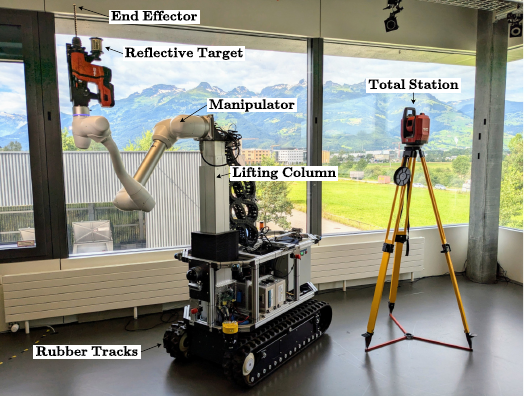}
\caption{A prototype of the Hilti JaiBot, together with a total station, used in this study to evaluate our model. The individual components of the system are annotated in the image.}
\label{fig:trailblazer}
\vspace{\capreductionfigures} % Adjust negative value as needed
\end{figure}

In environments where closed-loop positioning is infeasible, stationing routines offer a way to overcome this limitation~\cite{ercan_automated_2019,jaibot_patent}. While keeping the robot's base fixed, the robot's exact position and orientation are measured at the start of operation using a total station. With this initial information, the robot relies only on internal models to estimate movement, without needing ongoing measurements from the total station. However, in the absence of continuous tracking, errors can accumulate over the manipulator's motion due to uncertainties in the kinematic chain, such as mechanical flexing or joint looseness. 

In this context, we aim to address two research questions: How can we improve the positioning accuracy of long kinematic chains under varying environmental conditions, and, how can we adapt a deflection model to tightly integrate additional sensing on parts of the model expected to change?

To explore these questions, we present a method to significantly increase the accuracy of stationing methods by combining a deflection and backlash model of the lifting column with multiple high-accuracy accelerometers to measure the (biased) direction of gravity. We tested our proposed algorithm on various datasets under different environmental conditions, comparing it against traditional and model-based stationing methods, demonstrating superior performance. The key contributions of this work are:

\begin{itemize}
\item A modular framework based on a factor graph formulation to estimate the state of the kinematic chain and calibrate the full robotic model.
\item A physics-oriented approach to fusing backlash and deflection models of multiple joints with high-accuracy accelerometer data, implemented on a large construction robot.
\item A publicly released dataset of 3090 accurate end effector and accelerometer measurements used for extensive accuracy tests of the proposed model.
\end{itemize}
To the best of our knowledge, this is the first work to model the deflection of a robot's kinematic chain using a factor graph formulation. This approach leverages the modularity and scalability of factor graphs to efficiently represent and solve complex, nonlinear, and probabilistic models.

\section{RELATED WORK}\label{sec:related_work}

Positioning the end effector of a robot is a well-known challenge in construction robotics which spans a variety of research fields~\cite{ardiny_construction_2015}. A large focus is put on localizing directly against the building model to eliminate the need of a total station by using onboard LiDAR sensors. One method is to identify common features such as lines~\cite{hendrikx_connecting_2021} or corners~\cite{khoa_le_quenda-bot_2023} in 2D. Given a good initial guess, localization can also be performed directly against the surface data in 3D by converting the model into a mesh. To further increase accuracy, the matching can be executed on selected walls only relevant to the local task~\cite{blum_precise_2020}. The deviations between the measurements and the building model can be used to update the model to the as-built state~\cite{ercan_jenny_online_2020}. As off-the-self LiDAR sensors have large noise, some researchers build their own sparse LiDAR using multiple mm-accurate electronic range meters~\cite{gawel_fully-integrated_2019}.

Achieving robust mm-accurate positioning directly against the building remains an open challenge, especially under real-world conditions where a large gap exists between the as-planned and as-built state of buildings. Therefore, as of today, the use of total stations still is seen as state-of-the-art when it comes to mm-accurate localization in construction.

To achieve this level of accuracy when using stationing routines, however, an accurate kinematic model of the system is required. In robotic manipulation, the problem of non-rigidity of the kinematic chain of robot arms is a well-known phenomena. One method to identify the nonlinearities such as the motor backlash, stiffness and internal friction is to clamp the end effector to a world-fixed force-torque sensor and excite all degrees of freedom~\cite{lehmann_robot_2013}.

Most methods, however, rely on the identification using external measurement devices such as total stations. The most common method employed to model nonlinearities in the kinematic chain is the \ac{vjm} which adds virtual springs in the chain to model joint and link deflections~\cite{wang_model-based_2022,lim_practical_nodate,zhou_simultaneous_2014,cho_real-time_2007}. More recently, the combination of such models with neural networks was studied with the aim of compensating unmodeled effects using a data-driven approach~\cite{nguyen_new_2019}.

Fully relying on a calibrated model only works in environments where the calibration is not expected to change significantly. In rough environments and on mobile manipulators where dirt, temperature and vibrations due to base motion or the task execution such as the drilling process are expected to have significant long-term impact, adding more sensors along the kinematic chain helps to preserve accuracy over time. Adding an IMU to the end effector, for example, allows for online refinement of the kinematic parameters~\cite{du_imu-based_2013}. Similarly, using a filtering approach to fuse forward kinematics and accelerometer readings allows for smoothed pose estimation of the end effector~\cite{chen_pose_2015}.

Unless the deflection and backlash of the system are explicitly modeled, fusion of IMU and forward kinematics will only lead to partial improvement in accuracy as the measurement model still being based on an inaccurate rigid-body model. A simple way to incorporate the IMU measurements is to utilize them to estimate joint states of the system which are fed into a deflection model~\cite{regina_lauer_state_2023}.

This approach assumes the measurement being independent from the deflection, therefore ignoring the important aspect of the measurement changing due to the deflection. One method to partly counteract this is to use a filtering scheme like a complementary filter to get a smoothed estimate between a model, such as the tilt of a robot base calculated from displacement sensors of the suspension, and the acceleration measurements~\cite{feng_measurement_2023}.

Unless the bias of the accelerometer is modeled, the accuracy of the system's attitude estimate will be significantly compromised. Additionally, fusing the measurements with a model in a loosely-coupled manner, such as with a complementary filter, introduces additional errors. This occurs because the complementary filter tends to oversimplify the measurement models and the correlations between sensors and models, leading to the fusion of information that may be redundant or not optimally utilized.

In this publication, we address these issues by tightly coupling accelerometer measurements with a deflection model of the system. We formulate the model exploiting physical properties to ensure that the fusion process leverages the unique contributions of each sensor, rather than merging overlapping or redundant data. This approach allows for a more accurate and reliable estimation by utilizing the specific strengths of each measurement type in relation to the state being estimated, providing robustness against model changes.

\section{PROBLEM FORMULATION}\label{sec:problem_formulation}
In this paper, we address the problem of defining a calibration and stationing method for the system described in \Cref{sec:hardware} which considers and minimizes the effects of the uncertainties described in \Cref{sec:sources_of_uncertainty} on the final accuracy.

\subsection{System Setup}\label{sec:hardware}
We validate our approach on the construction robot depicted in \Cref{fig:trailblazer}, however, we highlight that the proposed framework is modular and can be applied to a variety of robotic systems.
The robot weights about $700\si{kg}$ and is composed of a base with rubber tracks, an extendable lifting column, a Doosan manipulator, and a drilling end effector. The column has two segments actuated by a single motor, ensuring both extend simultaneously by the same amount. The system is equipped with encoders to measure the joint states of the lifting column and the manipulator. At the end effector, a reflective target, which can be tracked with a total station, is attached. Additionally, two ADXL355\footnote{ADXL355 Product Page: \url{https://analog.com/en/products/adxl355.html}} high-accuracy accelerometers are mounted on the robot, one on the rigid frame of the base platform and one on the tip of the lifting column. We refer to \Cref{fig:deflection_model} for an illustration and \Cref{tab:allan_adxl355} for the results of an Allan Variance calibration.

\begin{table}[t]
    \vspace{\additionaltopvspacetables}
    \centering
    \scriptsize
    \begin{NiceTabularX}{\linewidth}{c|*{2}{C}}
\hline
\textbf{Axis} & \textbf{Noise Density $\left[\frac{m}{s^2}\frac{1}{\sqrt{Hz}}\right]$} & \textbf{Random Walk $\left[\frac{m}{s^3}\frac{1}{\sqrt{Hz}}\right]$} \\ \hline
x             & 9.645e-5                                                       & 1.794e-6                                                     \\ \hline
y             & 9.759e-5                                                       & 2.459e-6                                                     \\ \hline
z             & 1.387e-4                                                       & 2.515e-6                                                     \\ \hline

\end{NiceTabularX}

    \caption{Noise model of the ADXL355 from an Allan Variance identification.}
    \label{tab:allan_adxl355}
    \vspace{\capreductiontables} % Adjust negative value as needed
\end{table}

\subsection{Sources of Uncertainty}\label{sec:sources_of_uncertainty}
The system presented in the previous section is subject to several sources of uncertainty that make the real kinematic chain deviate from a rigid model:
\begin{itemize}
\item The rubber tracks compress under load. This behavior is highly dependent on the shape of the floor beneath the robot.
\item The joints between the lifting column segments compress under load as the segments have hard rubber bearings to allow smooth extension of the column.
\item The joints between the lifting column segments are affected by backlash due to gaps in the bearings.
\end{itemize}

Relying solely on a fixed model to capture these effects does not lead to acceptable results due to several reasons. First, the shape and properties of the floor the robot is standing on changes the deflective behavior of the rubber tracks. Second, variations in temperature have a large impact on the deflective behavior of both the rubber tracks and the column joints. Third, abrasion of the hard rubber bearings of the column joints increases their backlash over time.

\section{METHODOLOGY}\label{sec:method}
\newcommand*{\measurement}[1]{\pmb{\mathcal{M}}_\text{#1}}

In this section, we first describe the stationing routine at a high level (\Cref{sec:stationing_routine}) before diving into the associated factor graph formulation that allows us to perform state estimation (\Cref{sec:factor_graph}). We describe how we model and incorporate deflective behavior (\Cref{sec:hyb_model_facs}) and measurements (\Cref{sec:meas_factors}) into the factor graph and which physical properties allow to further constrain the optimization (\Cref{sec:constraining_models}).

In the following, we use small letters such as $a$ to represent single variables, small bold letters such as $\mathbf{a}$ to represent vectors and capital bold letters such as $\mathbf{A}$ to represent matrices. Specifically, the notation $\mathbf{T}_\text{A,B} \in \SEthree$ is used to represent transformation matrices transforming from frame B to A, with $\pospart{\mathbf{T}_\text{A,B}} = \mathbf{t}_\text{A,B} \in \Rthree$ and $\rotpart{\mathbf{T}_\text{A,B}} = \mathbf{R}_\text{A,B} \in \SOthree$ denoting the respective translational and rotational part. We use italic notation \textit{VarFactor} to denote a generic factor Var in the factor graph. We refer to the joint states and accelerometer measurements, which are available at all times, as internal measurements, denoted by $\measurement{in}$. In contrast, the total station measurements, which are only available when line-of-sight is present, are called external measurements, denoted by $\measurement{ex}$. We use the superscript $(\cdot)^{(t)}$ to denote the timestamp of a state, residual or measurement, if the respective object is added at a specific time instance.

\subsection{Stationing Routine}\label{sec:stationing_routine}
\newcommand*{\setof}[1]{\{#1\}}
\newcommand*{\setofwt}[2]{\setof{#1}^{(#2)}}
\newcommand*{\eepos}{_\text{world}\mathbf{p}_\text{ee}}
\newcommand*{\calibration}{\pmb{\mathcal{C}}}
\newcommand*{\jointstate}{\bm{\varphi}}
\newcommand*{\jointstates}[1]{\setofwt{\jointstate}{#1}}
\newcommand*{\measurements}[1]{\setofwt{\measurement{in}, \measurement{ex}}{#1}}

\begin{figure}[t]
\vspace{\topvspacealgorithm}
\begin{algorithm}[H]
\small{
\caption{Stationing Routine Calibration and Usage}\label{alg:stationing_routine}

\textbf{1. Calibration Procedure}
\begin{algorithmic}[1]
\State \textbf{define} $\jointstates{c}$ \Comment{Set of calibration joint configurations}
\State $\measurements{c} \gets \fmoveToAndRecordPoints(\jointstates{c})$ %\Comment{Record points}
\State $\calibration \gets \fcalibrate(\measurements{c})$ \Comment{Calibrate the model}
\State \textbf{return} $\calibration$ %\Comment{Transform to 3D Space}
\end{algorithmic}

\vspace{0.5em}
\textbf{2. Stationing Routine}
\begin{algorithmic}[1]
\State \textbf{define} $\jointstates{s}$ \Comment{Set of stationing joint configurations}
\State \textbf{define} $\jointstates{e}$ \Comment{Set of evaluation joint configurations}
% 2.1 Recording Stationing Points
\Statex \!\!\!\textbf{2.1 Recording Stationing Points}
\State $\measurements{s} \gets \fmoveToAndRecordPoints(\jointstates{s})$ % \Comment{Record points}

% 2.2 Extrapolation and Task Execution
\Statex \!\!\!\textbf{2.2 Extrapolation and Task Execution}
\For{$\jointstate \in \jointstates{e}$} \Comment{Iterate over evaluation configurations}
    \State $\measurement{in} \gets \fmoveToAndRecordPoint(\jointstate)$ %\Comment{Record evaluation point}
    \State $\eepos = \foptimize(\measurement{in} | \measurements{s}, \calibration)$ \Comment{Optimize}
    \State $\fexecuteTask(\eepos)$ \Comment{Execute task given position}
\EndFor
\end{algorithmic}
}
\end{algorithm}
\end{figure}

The structure of our stationing routine is outlined in \Cref{alg:stationing_routine}. We use three types of joint configurations and points, where the superscript
\begin{itemize}
    \item $(c)$ denotes calibration points recorded once in the factory (external measurements available), allowing to calibrate the system
    \item $(s)$ denotes stationing points recorded after motion of the mobile base (external measurements available), allowing to infer the global pose of the system
    \item $(e)$ denotes evaluation points where a task shall be executed (only internal measurements available), allowing to extrapolate the inferred global pose
\end{itemize}

Before using the routine, the robotic system needs to be calibrated by recording many internal and external measurements ($\measurements{c}$) in different joint configurations $\jointstates{c}$ ($\fmoveToAndRecordPoints$). These measurements are then used to find the calibration parameters ($\calibration$) that are composed of the kinematic calibration of the robot arm and the associated deflection parameters ($\fcalibrate$).

The stationing routine consists of two phases: First, stationing points $\measurements{s}$ are recorded ($\fmoveToAndRecordPoints$). In a classical stationing routine, these points primarily serve as a way to identify where the robot is located w.r.t. a world-fixed frame, but are here also used to online calibrate the model. Then, the robot is put into a joint configuration of interest $\jointstate$ and an evaluation point ($\measurement{in}$) is recorded ($\fmoveToAndRecordPoint$). At this configuration, we want to accurately know the position of the end effector ($\eepos$) by extrapolating from the stationing points using the kinematic model. To achieve this, we jointly optimize the states of our model given the evaluation measurement, the stationing points, and the calibration parameters ($\foptimize$). Finally, given the accurate position of the end effector, the task is executed ($\fexecuteTask$) - for instance, this could be a local correction motion to position the end effector precisely at the desired position and drill a hole.

Note that a straight forward way would be to jointly optimize with all points (calibration, stationing and evaluation) while taking care of temporal effects such as a moved world frame. Due to the large number of calibration points, the optimization would take significantly longer, about \SI{3}{\second} (see \Cref{sec:results}). Therefore, we use a separate calibration procedure to obtain $\calibration$ and constrain the optimization for stationing. Also, note that for every evaluation joint configuration, we solve a new optimization problem. This is to avoid biasing the solution with other evaluation points and make it independent of the order of joint configurations.

\subsection{Factor Graph Formulation}\label{sec:factor_graph}
In this section, we present the implementation of the core functions $\fcalibrate$ and $\foptimize$ used in \Cref{alg:stationing_routine} to calibrate the model from calibration points and compute the end effector position from evaluation measurements, stationing measurements, and calibration parameters. Given the nonlinear nature of the system presented in \Cref{sec:problem_formulation}, such computation results in a nonlinear optimization problem. In this paper, we model the optimization using a factor graph, a bipartite graph which represents the factorization of a probability distribution. Each node represents a probability distribution of a variable being optimized for while the factors
represent probabilistic constraints. Factor graphs are often used in robotics\cite{nubert_graph-based_2022,buchanan_learning_2022,lanegger2023chasing} due to their ability to efficiently represent and solve complex, nonlinear, and probabilistic models in a modular and scalable manner.
We base our implementation on GTSAM~\cite{gtsam}, a library commonly used in robotics.

Given the state $\mathbf{x}$ that we want to estimate and measurements $\mathbf{z}$, the target is to perform maximum a posteriori inference on the factorized probability distribution to find the optimal set of states $\mathbf{x}^*$ fitting the measurements. Under the assumption of Gaussian noise distributions for measurements $p(\mathbf{z}_i | \mathbf{x}) \sim \exp ( -0.5 || \mathbf{r}_i(\mathbf{x}) ||_{\mathbf{\Sigma}_{z,i}^{-1}}^2)$ and Gaussian distributions for states $p(\mathbf{x}_j) \sim \exp ( -0.5 || \mathbf{x}_j - \boldsymbol{\mu}_j ||_{\mathbf{\Sigma}_{x,j}^{-1}}^2 )$, the problem results in a least-squares optimization
\begin{align}
\begin{split}
    \mathbf{x}^* &= \underset{\mathbf{x}}{\arg\max} \,\, p(\mathbf{x} | \mathbf{z}) = \underset{\mathbf{x}}{\arg\min} \log p(\mathbf{z} | \mathbf{x}) + \log p(\mathbf{x}) \\
    &= \underset{\mathbf{x}}{\arg\min} \sum_i  ||\mathbf{r}_i(\mathbf{x})||_{\mathbf{\Sigma}_{z,i}^{-1}}^2 + \sum_j  || \mathbf{x}_j - \boldsymbol{\mu}_j ||_{\mathbf{\Sigma}_{x,j}^{-1}}^2
\end{split}
\end{align}
with $\mathbf{\Sigma}_{z,i}, \mathbf{\Sigma}_{x,j}$ denoting the covariance matrix of the corresponding measurement and state respectively and $\mathbf{r}_i(\mathbf{x})$ denoting the measurement residual.

\begin{figure}[t]
    \vspace{\additionaltopvspacefigures}
    \centering
    \includegraphics[width=1.00\linewidth]{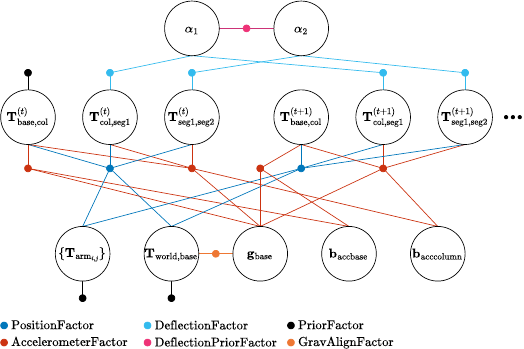}
    \caption{Factor graph formulation of the model. We use the \textit{DeflectionPriorFactor} only during stationing and the \textit{PositionFactor} only if $\measurement{ex}$ available.}
    \label{fig:factor_graph}
    \vspace{\capreductionfigures} % Adjust negative value as needed
\end{figure}

\Cref{fig:factor_graph} visualizes the factor graph used in this publication. The functions $\fcalibrate$ and $\foptimize$ correspond to the processes of optimizing the factor graph using the factors described in this section. These functions return the optimized states or the calculated end effector position from the corresponding optimized states, respectively. The state vector $\mathbf{x}$ consists of the current pose of the robot base w.r.t. the total station ($\mathbf{T}_\text{world,base} \in \SEthree$), the total station gravity alignment ($\mathbf{g}_\text{base} \in \Rthree$), the current biases of the accelerometers ($\mathbf{b}_\text{accbase},\mathbf{b}_\text{acccolumn} \in \Rthree$), the deflection coefficients ($\boldsymbol{\alpha}_1, \boldsymbol{\alpha}_2 \in \Rsix \geq \mathbf{0}$), the calibration of the robot arm ($\armcalibration \in \RNbyfourbyfour, \, \mathbf{T}_{\text{arm}_{i,j}} \in \SEthree$) and the state of the kinematic chain at each timestamp ($\mathbf{T}_{\text{base,col}}^{(t)}, \mathbf{T}_{\text{col,seg1}}^{(t)}, \mathbf{T}_{\text{seg1,seg2}}^{(t)} \in \SEthree$). By introducing the following constraining models as factors into the graph, the full state vector - including the biases of the accelerometers - become observable through the stationing and evaluation points.

\subsubsection{Deflection Models}\label{sec:hyb_model_facs}
\begin{figure}[t]
\vspace{\additionaltopvspacefigures}    
\includegraphics[width=1.00\linewidth]{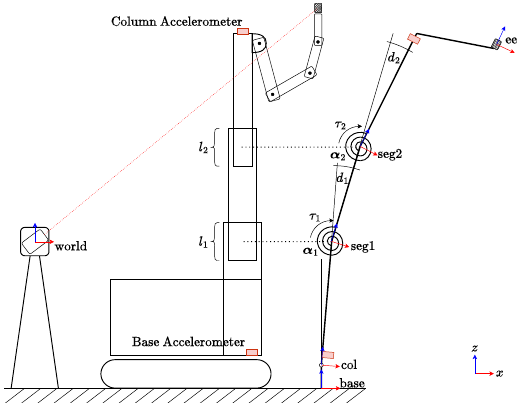}
\caption{A 2D visualization of the deflection model. The model is composed of a compliance part modeling the base-environment interaction and linear springs with backlash for each column joint.}
\label{fig:deflection_model}
    \vspace{\capreductionfigures} % Adjust negative value as needed
\end{figure}

As stated in \Cref{sec:sources_of_uncertainty}, the system is affected by multiple sources of uncertainty which we address as depicted in \Cref{fig:deflection_model}. 
We model rubber track deflection as a compliance state $\mathbf{T}_{\text{base,col}}^{(t)}$ in the kinematic chain, optimized at each timestamp. As shown in~\cite{feng_measurement_2023}, base tilting is observable via a base-mounted accelerometer.

The behavior of the lifting column is affected by both backlash and deflection. We model this behavior separately for each column joint in the \textit{DeflectionFactor} as a series of two linear springs~\cite{wang_model-based_2022,lim_practical_nodate,zhou_simultaneous_2014} for roll and pitch but additionally affected by backlash.
\begin{equation}\label{eq:deflection_model}
\begin{split}
\mathbf{r}_\text{defl}^{(t)}(\mathbf{T}_\text{parent,child}^{(t)}, \boldsymbol{\alpha}_r, \boldsymbol{\alpha}_p) =
\rotpart{\mathbf{T}_\text{parent,child}^{(t)}} \boxminus \mathbf{R}_\text{model}^{(t)} \\
\mathbf{R}_\text{model}^{(t)} = \mathbf{R}_x(d(\tau_x^{(t)}, l, \boldsymbol{\alpha}_r)) \mathbf{R}_y(d(\tau_y^{(t)}, l, \boldsymbol{\alpha}_p))\\
d(\tau, l, \boldsymbol{\alpha}) = \frac{\alpha_1}{2l} \arctan(\alpha_2 \tau) + \frac{\alpha_0}{l \tau}
\end{split}
\end{equation}
with $\mathbf{T}_\text{parent,child}^{(t)}$ denoting the transform at time $t$ which is being optimized for, $\boldsymbol{\alpha}_r$ and $\boldsymbol{\alpha}_p$ the deflection parameters for roll and pitch axis respectively and $\mathbf{R}_\text{model}^{(t)}$ the expected rotation according to our deflection model.
The $\boxminus$ operator, $\mathbf{R}_1 \boxminus \mathbf{R}_2 = \log(\mathbf{R}_1^T \mathbf{R}_2)^\vee$, maps the difference between two rotations in $\SOthree$ to the Lie algebra $\mathfrak{so}(3)$ using the logarithm map and expresses it in the vector space $\Rthree$ using the vee operator $(\cdot)^\vee$~\cite{sola2018micro}.
The deflection is estimated as two angular springs in series, each with backlash and deflection, depending on the segment overlap $l$ and the torques $\tau_x^{(t)}, \tau_y^{(t)}$ for the respective axis. Simply speaking, $\alpha_0$ models the spring compliance, $\alpha_1$ the magnitude of the backlash, and $\alpha_2$ the steepness of the continuous backlash approximation. 

These parameters, however, change over time due to dependencies on factors like temperature and wear. To tightly integrate accelerometer data only on the parts of the model expected to change, we re-optimize the parameters during stationing while constraining them with the \textit{DeflectionPriorFactor}. The idea behind this constraint is as follows: 
\begin{itemize}
    \item We expect that the deflection coefficient $\alpha_0$, due to a change in temperature, changes equally with a multiplicative factor for all joints of the column and for each rotation axis separately.
    \item We expect that the backlash coefficient $\alpha_1$, due to abrasion over time, changes equally with an additive factor for all joints of the column and for each rotation axis separately.
\end{itemize}
This constraint allows to make the full state of the lifting column observable from the accelerometer on the tip of the column. Mathematically, the factor is formulated as
\begin{equation}
\begin{split}
\mathbf{r}_\text{deflprior}(\boldsymbol{\alpha}_l, \boldsymbol{\alpha}_u) =
\begin{bmatrix}
           \alpha_{l,0} \, \beta_{u,0} - \alpha_{u,0} \, \beta_{l,0} \\
           (\alpha_{l,1} - \beta_{l,1}) - (\alpha_{u,1} - \beta_{u,1})
\end{bmatrix}
\end{split}
\end{equation}
with $\boldsymbol{\alpha}_l, \boldsymbol{\alpha}_u$ denoting the deflection coefficients for the lower and upper joint of the lifting column and $\boldsymbol{\beta}_l, \boldsymbol{\beta}_u$ denoting the corresponding resulting coefficients received during the calibration of the system. To enforce these physical constraints, both the \textit{DeflectionFactor} and \textit{DeflectionPriorFactor} are modeled with hard constraints.

\subsubsection{Measurement Models}\label{sec:meas_factors}

Observability of the problem requires the incorporation of both the position measurements of the total station and acceleration measurements of the accelerometers. We relate the respective measurements to the state of the kinematic chain at its corresponding time stamp and the remaining states through the \textit{PositionFactor} and \textit{AccelerometerFactor}.

The \textit{PositionFactor} compares the estimated position of the reflector mounted at the end effector given the model against the actual measurement
\begin{equation}\label{eq:position_factor}
\begin{split}
&\mathbf{r}_\text{pos}^{(t)}(\mathbf{T}_\text{world,base}, \armcalibration, \kinematicchain) =\\
&\,\tilde{\mathbf{m}}_\text{pos}^{(t)} - \pospart{\mathbf{T}_\text{world,base} \left( \prodoverkinchain \right) \text{fk}(t, \armcalibration) }
\end{split}
\end{equation}
with $\tilde{\mathbf{m}}_\text{pos}^{(t)}$ denoting the total station measurement, $\kinematicchain$ denoting the ordered set of kinematic chain transforms from the base to the robot arm (in Figure~\ref{fig:factor_graph} $\{\mathbf{T}_\text{base,col}^{(t)}, \mathbf{T}_\text{col,seg1}^{(t)}, \mathbf{T}_\text{seg1,seg2}^{(t)}\}$) and $\text{fk}(t, \armcalibration)$ denoting the forward kinematic transform of the robot arm at time $t$. Note that $\kinematicchain$ could also capture the deflection of the manipulator. In our experiments, however, we have only observed a comparably small improvement as the majority of deflection occurs in the lifting column.

Similarly to \Cref{eq:position_factor}, the \textit{AccelerationFactor} compares the estimated acceleration measurement of the accelerometers given the model against the actual measurement
\begin{equation}
\begin{split}
&\mathbf{r}_\text{acc}^{(t)}(\mathbf{g}_\text{base}, \mathbf{b}, \kinematicchain) =\\
&\qquad\qquad\qquad\tilde{\mathbf{m}}_\text{acc}^{(t)} \!-\! \left( \rotpart{\prodoverkinchain} \!\!\!\!\!\textbf{g}_\text{base} \!+\! \textbf{b} \right)
\end{split}
\end{equation}
with $\tilde{\mathbf{m}}_\text{acc}^{(t)}$ denoting the acceleration measurement, $\mathbf{b}$ the corresponding sensor bias and $\kinematicchain$ denoting the ordered set of kinematic chain transforms from the base to the sensor (in Figure~\ref{fig:factor_graph}, the set for the base accelerometer is $\{\mathbf{T}_\text{base,col}^{(t)}, \mathbf{T}_\text{col,accbase}\}$, and for the accelerometer on the column, it is  $\{\mathbf{T}_\text{base,col}^{(t)}, \mathbf{T}_\text{col,seg1}^{(t)}, \mathbf{T}_\text{seg1,seg2}^{(t)}, \mathbf{T}_\text{seg2,acccolumn}\}$).

For each new joint configuration where we record an evaluation point, the corresponding states of the kinematic chain at the timestamp are added to the factor graph and an \textit{AccelerationFactor} is added and connected to them. For stationing or calibration points, additionally a \textit{PositionFactor} is spawned and connected. 

\subsubsection{Constraining Models}\label{sec:constraining_models}
In addition to the deflection (\Cref{sec:hyb_model_facs}) and measurement models (\Cref{sec:meas_factors}), there are more physical properties of the system which allow us to further constrain the optimization. 

The total station estimates the direction of gravity with high accuracy and aligns its reference frame so that its z-axis aligns with the estimated gravity vector. We capture this in the \textit{GravAlignFactor} by enforcing $\mathbf{g}_\text{base}$ to be z-axis-aligned with the world frame 
\begin{equation}
\begin{split}
\mathbf{r}_\text{grav}(\mathbf{T}_\text{world,base}, \mathbf{g}_\text{base}) =
\mathbf{g}_\text{world} - \rotpart{\mathbf{T}_\text{world,base}} \, \textbf{g}_\text{base}
\end{split}
\end{equation}
with $\mathbf{g}_\text{world} = [0, 0, 9.807]^T$ denoting the gravity vector.

We know that our robot is not able to tilt by more than a threshold - we include this as a \textit{PriorFactor} acting on $\mathbf{T}_\text{world,base}$ constraining the state to rotations about roll and pitch in a defined region, preventing upside-down solutions.

The base of the robot is mainly affected by tilting forward/backwards and left/right due to the expected deflection of the rubber tracks. We include this effect as a \textit{PriorFactor} acting on each $\mathbf{T}_{\text{base,col}}^{(t)}$, fixing the position and yaw to CAD values while only allowing rotations about roll and pitch within the expected region of deflection.

During system calibration, we aim to optimize the transforms of the robot arm, denoted as $\armcalibration$. Particularly when calibrating the arm with end effector poses that vary in fewer than six degrees of freedom, such as when focusing solely on ceiling drilling, multiple solutions may exist. During calibration, we therefore constrain the arm calibration with a \textit{PriorFactor} to an expected region of variability. During stationing, we freeze the acquired calibration using the \textit{PriorFactor} as a hard constraint.

\subsubsection{State Initialization}\label{sec:state_init}

To solve the optimization problem, we use the Levenberg-Marquardt optimizer, which is an iterative solver requiring an initial guess of the state vector. The current pose of the robot base w.r.t. the total station $(\mathbf{T}_\text{world,base})$ is initialized using the Kabsch algorithm~\cite{Kabsch1976} applied to the stationing points. This involves aligning the total station measurements with the prism's position obtained from forward kinematics using the rigid body model. 

The state of the kinematic chain at each timestamp $(\mathbf{T}_{\text{base,col}}^{(t)}, \mathbf{T}_{\text{col,seg1}}^{(t)}, \mathbf{T}_{\text{seg1,seg2}}^{(t)})$ is also initialized from the rigid body model. The total station gravity alignment $(\mathbf{g}_\text{base})$ and the current biases of the accelerometers $(\mathbf{b}_\text{accbase},\mathbf{b}_\text{acccolumn})$ are initialized to zero.

During system calibration, the deflection coefficients $(\boldsymbol{\alpha}_1, \boldsymbol{\alpha}_2)$ are initialized to zero, and the calibration of the robot arm $(\armcalibration)$ are set according to the rigid body model. For stationing, however, the deflection coefficients and the arm calibration parameters are initialized using the results from the calibration procedure.

\section{EXPERIMENTS}\label{sec:experiments}
To evaluate the effectiveness of our approach in improving the positioning accuracy of long kinematic chains, we deployed the proposed algorithm on datasets recorded with the construction robot shown in \Cref{fig:trailblazer}. These datasets capture a variety of environments, as described in \Cref{sec:datasets}. We compare the accuracy of our method against two baselines and perform an ablation study, detailed in \Cref{sec:results}.

\subsection{Datasets}\label{sec:datasets}
We captured seven distinct datasets under conditions simulating realistic construction site disturbances and released them publicly\footnote{Datasets available at:\newline\hspace*{1.5em}\url{https://github.com/Hilti-Research/construction-robot-stationing-datasets}}. Each dataset contains identical waypoints that the robot arm follows in a consistent grid pattern through 448 points, with variations only in the induced disturbances.

\begin{figure}[t]
\vspace{\additionaltopvspacefigures}
\includegraphics[width=\linewidth]{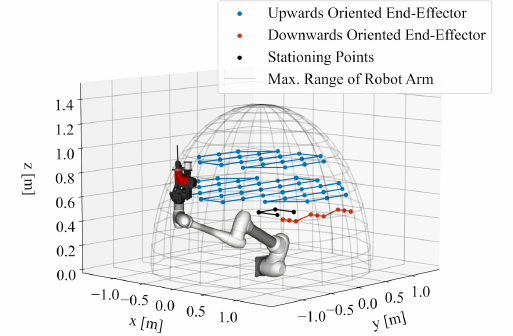}
\caption{Waypoints, through which the robot navigates for every lifting column height using the standard solution space.}
\label{fig:waypoints}
    \vspace{\capreductionfigures} % Adjust negative value as needed
\end{figure}

The parameters subject to changes during dataset recording are (i) the height of the lifting column, (ii) the height of the robot arm, (iii) the solution space of the robot arm and (iv) the vertical orientation of the end effector. The robot arm navigates through different layers of points, categorized into stationing points, and points where the arm has an upwards or a downwards end effector configuration, as illustrated in \Cref{fig:waypoints}. This process is done for two different solution spaces of the manipulator and for every lifting column height. 

The following provides an overview of the datasets employed in this study.
Unless otherwise stated, all of the datasets were recorded at \SI{21}{\degreeCelsius} room temperature.

\textbf{Calibration Dataset (\dnFlat)}: Operation on level surface at room temperature with no external distrubances.

\begin{figure*}[t]
    \vspace{\additionaltopvspacefigures}
    \centering
    \begin{subfigure}{0.24\textwidth}
        \centering
        \includegraphics[clip, height=2.0cm]{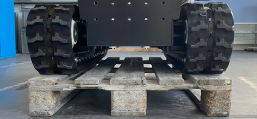}
        \caption{Pallet}
        \label{fig:pallet_front}
    \end{subfigure}
    \hfil
    \begin{subfigure}{0.24\textwidth}
        \centering
        \includegraphics[clip, height=2.0cm]{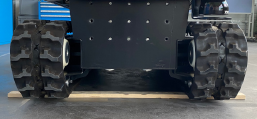}
        \caption{Orthogonal Wood}
        \label{fig:orthogonal_wood_front}
    \end{subfigure}
    \hfil
    \begin{subfigure}{0.24\textwidth}
        \centering
        \includegraphics[clip, height=2.0cm]{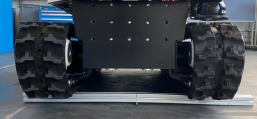}
        \caption{Seesaw}
        \label{fig:seesaw_front}
    \end{subfigure}    
    \hfil
    \begin{subfigure}{0.24\textwidth}
        \centering
        \includegraphics[clip, height=2.0cm]{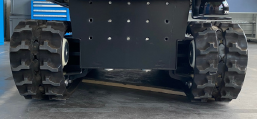}
        \caption{Diagonal Wood}
        \label{fig:diagonal_wood_front}
    \end{subfigure}
    \hfil
    \caption{Setups of the robot and its environment for the respective datasets.}
    \label{fig:robot_setup}
    \vspace{\capreductionfigures} % Adjust negative value as needed
\end{figure*}

\textbf{\dnPallet}: Robot positioned on a wooden pallet (\Cref{fig:pallet_front}).

\textbf{\dnOrthoWood}: Wooden slab orthogonal to robot's axis under front of tracks, causing a backward tilt (\Cref{fig:orthogonal_wood_front}).

\textbf{\dnLeftTrackWood}: Wooden slab under front of left track, causing a tilt backward and to the right.

\textbf{\dnSeesaw}: Front of tracks on aluminum seesaw, causing a strong tilt backward (\Cref{fig:seesaw_front}).

\textbf{\dnDiagWood}: Wooden slab under tracks, oriented diagonally, tilting robot backward and to left (\Cref{fig:diagonal_wood_front}).

\textbf{\dnOutdoor}: Captured outdoors at \SI{9}{\degreeCelsius}, contrasting with the \SI{21}{\degreeCelsius} of other datasets. Flat floor.
% \end{list}

\begin{figure}[t]
    \centering
    \includegraphics[width=\linewidth]{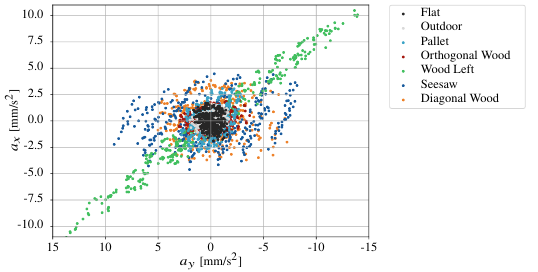}
    \caption{The centered distributions of the base accelerometer measurements in xy-plane for all datasets.}
    \label{fig:base_acceleration}
    \vspace{\capreductionfigures} % Adjust negative value as needed
\end{figure}

The disturbances in the datasets impact the deflective behavior of the system. We can see this effect in the visualization of the base accelerometer measurements in \Cref{fig:base_acceleration}. For example, in the \dnSeesaw~dataset, a constant backward tilt from the robot standing on an aluminium profile results in a rather stable x-axis acceleration. Conversely, the seesawing effect causes fluctuating y-axis values due to varying left/right tilts. The \dnLeftTrackWood~dataset exhibits the greatest tilt due to an unstable configuration around the robot's x-axis. This instability leads to proportional changes in both the x- and y-axis accelerations, appearing as a diagonal line in the figure.

\subsection{Results}\label{sec:results}
We evaluate the accuracy of our algorithm on a variety of datasets described in \Cref{sec:datasets}. Initially, the system is calibrated on a flat dataset. We then assess the model's accuracy by analyzing how well it extrapolates from stationing with four stationing points (2.2 in \Cref{alg:stationing_routine}). For each evaluation point, we record the end effector's position using the total station, which serves as the ground truth, and compare it against the model's predicted position. It is important to note that on real construction sites, using the total station to record the end effector position at evaluation points is infeasible due to line-of-sight constraints.

We use two baselines to evaluate our proposed solution \ac{fg}. First, we utilize the \ac{vjm}, with the springs positioned at the same location as in our model, as shown in \Cref{fig:deflection_model}, and parameters calibrated on the flat dataset and kept fixed throughout all experiments. Second, we introduce a simple yet novel approach by combining \ac{vjm} with base accelerometer data as a tilt sensor, which we denote as \ac{vjmbt}. In this approach, the tilt of the robot is calculated directly from the accelerometer measurements, assuming zero bias. Although we do not explicitly solve for the bias, the base accelerometer data is only used to capture relative tilts. The constant tilt offset introduced by any bias is inherently accounted for within $(\mathbf{T}_\text{world,base})$.

\begin{table*}[t]
\vspace{\additionaltopvspacetables}
    \centering
    % \tiny
    \scriptsize
    \begin{NiceTabularX}{\textwidth}{c|*{6}{C}|C}
\hline
 & \textbf{Pallet} & \textbf{Orthogonal Wood} & \textbf{Wood Left} & \textbf{Seesaw} & \textbf{Diagonal Wood} & \textbf{Outdoor} & \textbf{Overall}\\
\hline
\tableconfigVJM & $4.7(11.0)|2.7(4.7)$ & $4.8(9.5)|2.6(4.2)$ & $9.6(14.7)|3.4(4.5)$ & $8.8(13.9)|3.0(4.0)$ & $5.4(10.3)|2.8(4.4)$ & $4.6(7.2)|3.1(5.2)$ & $7.0(14.7)|3.0(5.2)$\\
\tableconfigVJMBA & $4.4(10.0)|\mathbf{2.7}(4.7)$ & $4.6(9.5)|\mathbf{2.5}(4.2)$ & $4.3(9.5)|\mathbf{2.7}(4.2)$ & $6.8(12.6)|2.8(3.8)$ & $4.6(9.1)|2.6(4.3)$ & $4.5(7.3)|\mathbf{3.0}(5.3)$ & $5.1(12.6)|\mathbf{2.8}(5.3)$\\
\tableconfigFULL & $\mathbf{3.5}(6.2)|2.9(4.3)$ & $\mathbf{3.6}(4.6)|2.8(4.4)$ & $\mathbf{3.3}(4.5)|2.9(3.8)$ & $\mathbf{3.8}(6.2)|\mathbf{2.8}(4.2)$ & $\mathbf{3.0}(5.7)|\mathbf{2.4}(3.9)$ & $\mathbf{3.8}(4.9)|3.1(4.9)$ & $\mathbf{3.5}(6.2)|2.8(4.9)$\\
\hline

    \end{NiceTabularX}
    \caption{$95\%$ error threshold and maximum error in the xy-plane and z-axis in millimeters for different configurations and datasets, denoted as $\text{R95}_{xy}(\max_{xy})|\text{R95}_z(\max_z)$.}
    \label{table:res_table_indivstationing_square}
    \vspace{\capreductiontables} % Adjust negative value as needed
\end{table*}

\begin{table}[t]
    \centering
    \scriptsize
    \begin{NiceTabularX}{\linewidth}{c|*{4}{C}}
\hline
 & $\textbf{R95}_\textbf{xy}$ & $\textbf{max}_\textbf{xy}$ & $\textbf{R95}_\textbf{z}$ & $\textbf{max}_\textbf{z}$ \\
\hline
Full Model (\ac{fg})    & 3.5              & 6.2              & 2.8             & 4.9             \\ 
Without Base Accelerometer   & 4.1              & 8.1              & 2.4             & 4.4             \\ 
Without Column Accelerometer & 5.1              & 15.9             & 3.4             & 5.4             \\ 
Without Both Accelerometers  & 6.8              & 16.9             & 2.9             & 5.0             \\ \hline

\end{NiceTabularX}

\caption{Ablation study of our proposed approach \ac{fg} on the overall performance. All units in millimeters.}
    \label{table:ablation_study}
    \vspace{\capreductiontables} % Adjust negative value as needed    
\end{table}

\begin{figure}[ht]
    \centering
    \includegraphics[width=\linewidth]{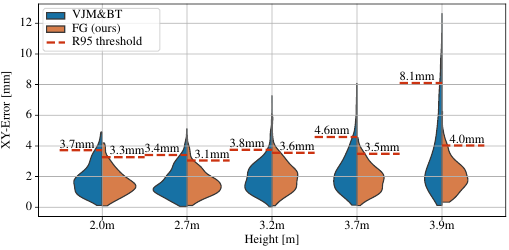}
    \caption{Histograms of XY-Errors for all individual column heights for our proposed approach \ac{fg} and the baseline \ac{vjmbt}.}
    \label{fig:matrix_heights_distros}
    \vspace{\capreductionfigures} % Adjust negative value as needed
\end{figure}

The calibration routine for our approach took about $\SI{3}{\second}$, while the setup and optimization of the factor graph for stationing at each evaluation point took on average $\SI{172}{\milli\second}$ on our Intel i7-1260P CPU. The resulting $95\%$ thresholds and maximum errors in the xy-plane as well as the z-axis are shown in \Cref{table:res_table_indivstationing_square}. We observe that our method consistently outperforms the baselines in xy-accuracy, effectively reducing the overall error by \percdiff{3.5}{7.0} compared to the state-of-the-art model and by \percdiff{3.5}{5.1} when incorporating the base tilt sensor in the baseline.

The most prominent improvement over \ac{vjm} is observed in \textit{\dnLeftTrackWood}, which, as stated in \Cref{sec:datasets}, features the most significant tilting of the base - something that \ac{vjm} is unable to capture. Similarly, the greatest improvement over \ac{vjmbt} is seen in \textit{\dnSeesaw}, where the deflection is nonlinear. While \ac{vjmbt} relies on a linear model calibrated on a flat dataset, our approach benefits from the column accelerometer, enabling it to account for the change in nonlinearity.

To better understand the source of our method’s improvements, we conducted an ablation study across all datasets, with the results shown in \Cref{table:ablation_study}. The largest drop in accuracy occurs when both accelerometers are removed, which is expected, as the model cannot predict base tilting, which is dependent on the environment. We also observe that if forced to choose between the base or column accelerometer, the column accelerometer offers better results, as it captures both the unmodelable base tilt and the partially modelable column deflection. However, only when using both accelerometers can the model efficiently distinguish between the two sources of deflection and account for both.

Although our model significantly improves accuracy in the xy-plane, it is less effective at reducing z-error. In fact, the ablation study shows that the z-error improves when the base accelerometer is removed. A likely explanation is that we model ground contact as a fixed point of rotation on the floor. However, analyzing the residuals in 3D suggests that the tracks act more like a four-bar linkage, introducing both rotational and translational motion at the contact point.

Finally, we investigate the column height dependency of the proposed approach. \Cref{fig:matrix_heights_distros} compares xy-error distributions across all datasets for different column heights between our approach and \ac{vjmbt}. At lower column heights, our approach performs on par with the baseline. However, as column height increases, the overlap between column segments decreases (\Cref{eq:deflection_model}), leading to larger deflections. By incorporating the column accelerometer, even with a linearized model, we can significantly better approximate the nonlinear nature of the deflection at higher column heights.

\section{CONCLUSION}\label{sec:conclusion}
In this work, we introduced a novel approach for robotic stationing aimed at enhancing the positioning accuracy of robotic systems in construction environments. By integrating a physically informed deflection model with high-precision accelerometer data, we demonstrated a significant improvement in xy-accuracy - achieving a $50\%$ reduction in error compared to the traditional approach, and a $31\%$ improvement over the baseline incorporating the base accelerometer - across a variety of datasets. These results highlight the effectiveness of our method in accounting for deflections and environmental factors that traditional models fail to capture adequately.

Our experiments, conducted across a diverse set of datasets capturing real-world construction disturbances, validated the robustness and adaptability of our model. The incorporation of accelerometer data further refined our deflection model, enabling it to capture the nuances of both base and column deflections that significantly impact accuracy. 
%In future work, we want to explore methods to capture backlash along the kinematic chain and effects at contact points using data-driven approaches.

\bibliographystyle{IEEEtran}
\bibliography{references} 

\end{document}